\pdfoutput=1

\documentclass[11pt]{article}

\usepackage[]{emnlp2021}

\usepackage{times}
\usepackage{latexsym}

\usepackage[T1]{fontenc}

\usepackage[utf8]{inputenc}

\usepackage{microtype}

\usepackage[utf8]{inputenc} 
\usepackage[T1]{fontenc}    
\usepackage{hyperref}       
\usepackage{url}            
\usepackage{booktabs}       
\usepackage{amsfonts}       
\usepackage{nicefrac}       
\usepackage{microtype}      
\usepackage{xcolor}
\usepackage{color}
\usepackage{colortbl}
\usepackage[colorlinks]{}
\usepackage{wrapfig,lipsum,booktabs}
\usepackage{listings}
\usepackage[colorlinks]{}
\usepackage{subcaption}
\usepackage{graphicx}
\usepackage{enumitem}

\title{Athena 2.0: Contextualized Dialogue Management \\ for an Alexa Prize SocialBot}

\author{Juraj Juraska, Kevin K. Bowden, Lena Reed, Vrindavan Harrison,    \\
{\bf Wen Cui, Omkar Patil, Rishi Rajasekaran, Angela Ramirez, Cecilia Li, } \\
	{\bf  Eduardo Zamora, Phillip Lee,  
	Jeshwanth Bheemanpally, Rohan Pandey,} \\
	{\bf Adwait Ratnaparkhi} and {\bf Marilyn Walker} \\
	University of California Santa Cruz, Santa Cruz, CA  \\
	\texttt{ jjuraska, kkbowden, lireed, vharriso, wcui7, opatil,} \\ 
		\texttt{  rrajasek, aramir62, yli331, ezamora9, plee35, jbheeman,} \\ 	\texttt{rspandey, adratnap, mawalker@ucsc.edu}
     }

\begin{document}
\maketitle
\begin{abstract}
Athena 2.0 is an Alexa Prize SocialBot that has been a finalist in the last two Alexa Prize Grand Challenges. One reason for Athena's success is its novel dialogue management strategy, which allows it to dynamically construct dialogues and responses from component modules, leading to novel conversations with every interaction. Here we describe Athena's system design and performance in the Alexa Prize during the 
20/21 competition. A live demo of Athena as well as video recordings  will provoke discussion on the state of the art in conversational AI.
\end{abstract}


\section{Introduction}

There has been tremendous progress over the last 10 years on  conversational AI, and a number of practical systems have been deployed. The Alexa Prize competition seeks to stimulate research and development on conversational AI for open-domain topic-oriented dialogue \cite{fang2018sounding18,liang2020gunrock,finch2020emora,harrison2020athena,alquist,curry2018alana}. However, the longstanding tension between hand-scripting the dialogue interaction, and producing systems that scale to new domains and types of interaction still remains \cite{Ericetal17,cervone2019natural}
Neural end-to-end spoken dialogue systems are not yet at a point where they perform well in interactions with real users \cite{chirpycardinal,gopalakrishnan2020neural,dinan2019build}.  

\begin{figure*}[t!hb]
    \centering
    \includegraphics[width=6.2in]{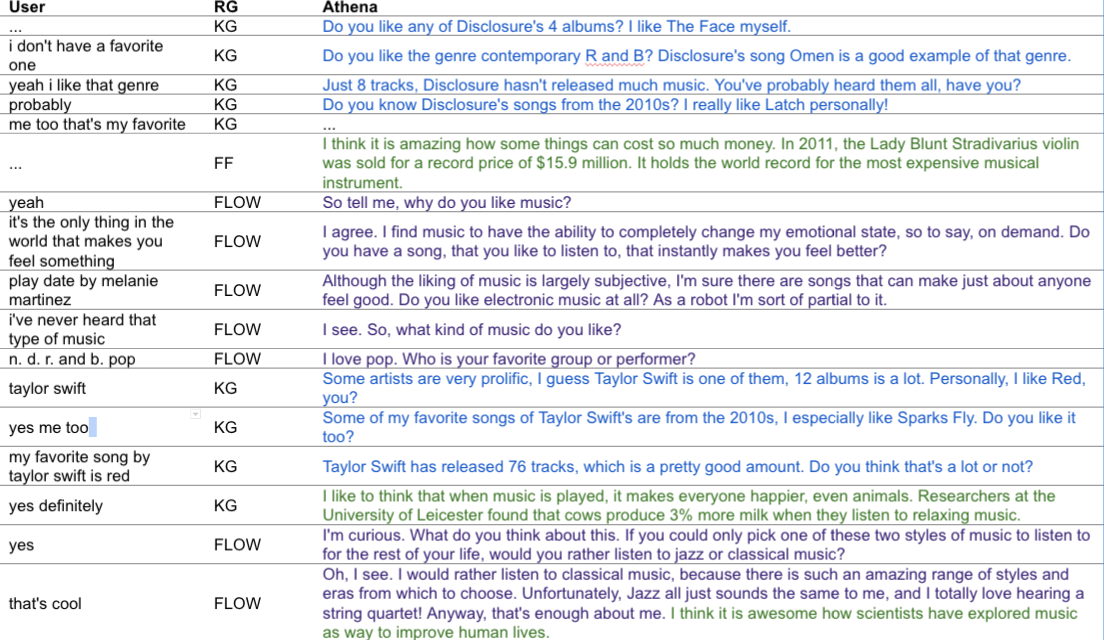}
    \caption{A subdialogue in which Athena discusses music with the user, interweaving the responses of 3 different RGs. Responses from the KG-based RG are highlighted in blue, responses from the entity-based indexing RG (fun facts) in green, and those from the flow-based RG in purple.}
    \label{fig:rg-interweaving-example}
    \vspace{-0.2cm}
\end{figure*}

Athena's dialogue management architecture aims to be scalable and dynamic, by supporting many different interactions for every topic, and by constructing system utterances by concatenating multiple dialogue acts that achieve different purposes \cite{Stent00}. A key aspect of Athena  is the existence of multiple Response Generators (RGs) for each topic, which can be flexibly interleaved during a particular interaction, as illustrated   in Figure~\ref{fig:rg-interweaving-example}.\footnote{In accordance with the Alexa Prize  rules on user privacy, these conversations are between team members and Athena. }
This approach contrasts with the commonly used approach of handcrafting  conversation flow-graphs for each topic,  a static directed graph where the nodes are the system utterances, and outgoing  edges are represent possible user replies. This approach has not changed for over 20 years \cite{seneff1998galaxy,Glass01,buntshuchetal98}, and its strengths and limitations are well known. Flow-graphs are hand constructed and thus  do not scale well. However, each system response can assume a fixed prior context, which allows it to support fluent and coherent dialogues with sufficient handcrafting. 

In contrast, the ability of Athena's Dialogue Manager (DM) to interleave RGs allows Athena to dynamically construct
conversations that never follow the same path. However, this more  flexible approach requires RGs to pay the overhead cost of continuously adapting to the current context, as described in Section~\ref{sec:dm}. By eschewing a graph-based representation of dialogue state, Athena’s DM is flexible enough to use RG responses in contexts that were not planned out prior to the conversation starting, and that do not need to follow rigid guidelines. We believe this modular dialogue management approach promises to scale to deeper and richer conversations, while at the same time allow new conversational topics to be easily added to and integrated into the system.    

\section{Athena Architecture and Overview}
\label{sec:arch-sec}

\begin{figure*}[t]
  \includegraphics[width=0.95\textwidth]{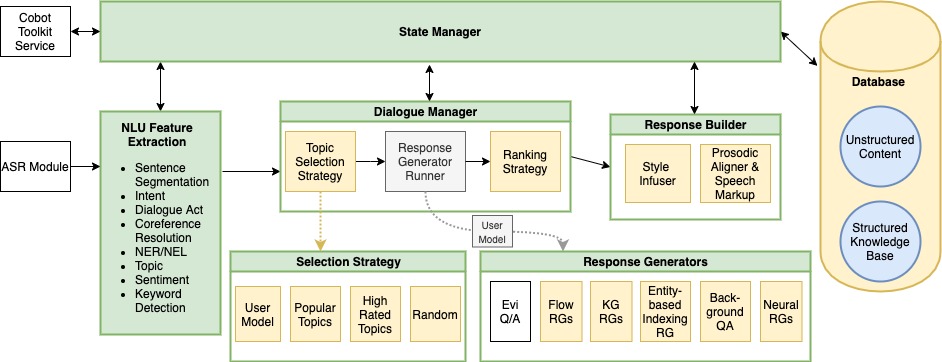}
  \caption{Athena's system architecture. 
  \label{fig:system-architecture}}
  \vspace{-.15in}
\end{figure*}

Figure~\ref{fig:system-architecture} details Athena's architecture. Athena is built using the Cobot framework provided by  Amazon \cite{khatri2018advancing}. It runs as an on-demand application that is initiated by an ``Alexa, let's chat'' user request to any Alexa-enabled device, such as an Amazon Echo or the Alexa app installed on a phone. During the Alexa Prize, Athena participates in about 9K conversations a week. The Cobot framework provides support for automatically scaling to large volumes of user traffic.

The inputs to Athena are the ASR hypotheses for a user's turn from Amazon, and a conversation ID that is used to retrieve the conversation history and state information from a back-end database. The ASR hypothesis is fed into a natural language understanding (NLU) pipeline that produces a set of NLU features for  the user utterance and conversation context. The NLU consists of  Cobot's module for topic classification, and Athena  modules
for utterance segmentation, dialogue act tagging, named entity recognition and linking, and coreference resolution \cite{harrison2020athena,Patiletal21}. The right-hand side of Figure~\ref{fig:system-architecture} indicates how Athena's RGs use knowledge bases and fun facts databases organized by topic and named entity. Athena uses  the Wikidata Knowledge Graph to aid in Named Entity Resolution and for Knowledge-Graph based RGs. These are essential for creating an intelligent and versatile conversational agent \cite{fang2018sounding18,chen2018gunrock}.

Based on the NLU features and conversation context, the Dialogue Manager (DM) calls specific Response Generators (RGs) to populate a response pool. 
The DM then applies a trained neural response ranker to select  from the response pool generated by the RGs. Finally, Athena's responses are spoken by Amazon's text-to-speech  service.

\section{Dialogue Management}
\label{sec:dm}

A Dialogue Manager (DM) for open-domain conversation faces a particularly challenging task  due to the universe of possible valid responses at each point of a conversation. While goal-oriented dialogues have a clear task completion objective which the DM can optimize when making decisions \cite{Walkeretal01,Walkeretal97,Walker00}, the DM for open-domain dialogues does not have an obvious way to measure the appropriateness of possible candidate responses.

\begin{figure}[thb]
  \centering
  \includegraphics[width=\columnwidth]{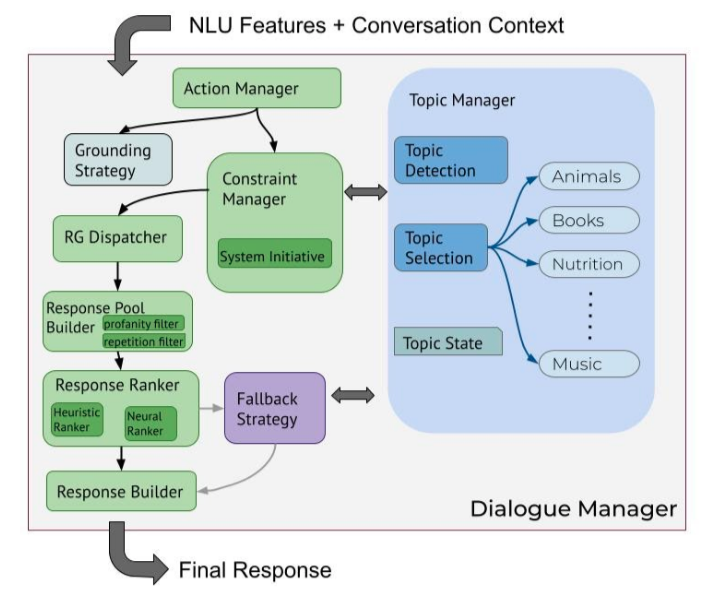}
  \vspace{-.25in}
  \caption{Dialogue manager architecture.}
  \label{fig:dm-architecture}
  \vspace{-.15in}
\end{figure}

Athena’s DM architecture can be decomposed into a number of sub-components, corresponding to phases of dialogue management,  oriented as a pipeline. The DM sub-modules in Figure~\ref{fig:dm-architecture} are described in
more detail in \citet{harrison2020athena}.

The Topic Manager in Figure~\ref{fig:dm-architecture} is responsible for classifying user utterances into topics, and the implementation of the DM's topic hierarchy. 
The topic hierarchy is a partially ordered list of topics in order of predicted ``goodness'' learned from  past conversations,
using a scoring function that combines  user ratings and the number of turns per
topic per conversation, as described in Section~\ref{eval-sec}. 
The topic hierarchy is a parameter for system-initiative topic initiations as well as suggesting topics for users to initiate. This makes it extremely easy to change which topics are promoted at any time, e.g., for collecting more
data on a particular topic. It can also be
personalized for each user. For example, if when asked about weekend activities, the user describes playing in a baseball league, we can prioritize talking about sports. This information persists across conversations. If the user is also an avid painter, but our system did not get a chance to discuss painting in the previous conversation, we will prioritize it when the user returns.

The interface between the DM and  the RGs in Figure~\ref{fig:dm-architecture},
is a contract-based approach. The DM passes a set of response conditions to the RGs, which the RGs must meet  for their response to be considered. This approach allows Athena  to have many RG types (see Section~\ref{sec:response-generation}).   


The Response Ranker is based on a BERT-based ranker fine-tuned on hand-annotated Alexa Prize conversation data \cite{Wolf2019HuggingFacesTS,Devlin_Chang_Lee_Toutanova_2018}.  
The current tuning set size is \textasciitilde 10K utterances. Annotation involves ranking candidate responses within a context of  five turns. We have repeatedly annotated additional data and retrained our response ranker, which is useful when, for example, new RGs are added to  Athena.


\section{Response Generation}
\label{sec:response-generation}

\begin{figure*}[t!hb]
    \vspace{-0.3cm}
    \centering
   \includegraphics[width=5.5in]{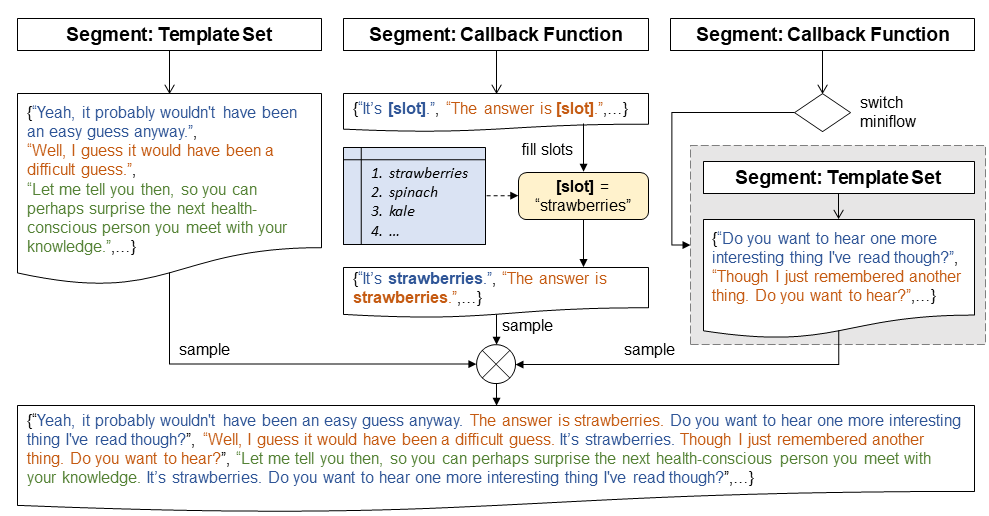}
    \vspace{-0.1cm}
    \caption{Illustration of response composition in Flow-RG.}
    \label{fig:flow-rg-response-composition}
    \vspace{-0.3cm}
\end{figure*}

Athena uses four types of RGs: Flow-RGs, Knowledge-Graph RGs, Entity-Based Indexing RGs, and Neural NLG RGs.

\subsection{Flow-RG}
\label{sec:flow-rg}

Flow-RG is a framework that we developed with the objective of creating robust and modular flow-based RGs. 
This is still the most reliable way to provide the DM with a pool of possible responses at each turn of the dialogue, even though such flows have to be handcrafted. Flow-based RGs exhibit context-awareness and fluency superior to other RG types, such as retrieval-based or neural. This RG design naturally has a rather limited support for user initiative, which we make up for with other RGs in Athena, and by ensuring the responses from different RGs get smoothly interwoven across multiple turns, as well as within a single turn.

An RG defined in this framework has three components. First, a \emph{flow graph} consisting of nodes specifying the responses, and edges determining which node of the flow to move on to given the current user utterance and dialogue state. Flow-RG enforces each next turn in the flow graph to be conditioned on the dialogue act(s) of the user utterance, while other features of the utterance -- such as its sentiment, or the presence of a named entity or a particular keyword -- are deemed secondary and are optional in branching conditions.\footnote{The other NLU features, such as named entities or anaphora, often only affect the choice of content in the next system response rather than its type/DA, and they do not have a value at each turn, unlike the DA feature.} This reduces the chance of Athena's subsequent response ignoring the user's intent, which can be anything from expressing an opinion, to requesting information, to merely acknowledging Athena's response in the previous turn.
The second component comprises \emph{response segment templates}, while the third component is a set of \emph{callback functions} that generate more context-dependent response segments.

A flow graph can be broken down into smaller \emph{miniflows} that are independent and can possibly be executed in an arbitrary order. Each RG then typically handles a single topic, with multiple miniflows being responsible for different subtopics. An example of multiple miniflows forming a cohesive dialogue can be seen in Appendix \ref{sec:appendix}.

\noindent{\bf Response Composition.}
The response in each turn is assembled from one or more \emph{segments} specified in the corresponding node. Each segment is defined either (1) in the form of a set of templates, or (2) as a callback function that returns a set of templates. While both offer an easy way to use paraphrases for increased diversity of the responses, the latter is more robust in that it can use the previous context and more of the NLU information about the user utterance.
Figure~\ref{fig:flow-rg-response-composition} shows the process of a response being assembled from three segments, two of which are different types of callback function: one fills a template slot with a value from the associated knowledge source, while the other initiates a new miniflow and composes the response text recursively, which ultimately corresponds to the last segment in the example.

When composing a response, each segment's final set of texts is sampled from, and all of them are concatenated. This is repeated until up to five different response candidates are composed. These are eventually all returned to the DM, which picks one of them that is not too similar to any of Athena's previous responses.

\noindent{\bf Interweaving with Other RGs.}
Every topic in Athena has a corresponding Flow-RG, and most
topics also have one or two other RGs that can interact with its Flow-RG
to dynamically construct a topical sub-dialogue. In line with the DM's way of response building, the final response in Flow-RG is split into three parts: an opener, a body, and a hand-off.
This response structure is particularly useful for creating seamless transitions whether it is between miniflows, or between two RGs. To this end, Flow-RG sets the response from an ending miniflow as the opener (typically, some form of acknowledgement of the user's response, or a short answer), and the body and hand-off parts are reserved for the intro response provided by a new miniflow.
The same mechanism is used for certain transitions from Flow-RG to a different RG, mainly: 
1) when the flow's content is exhausted, and transitions thus to a fallback response chosen by the DM that initiates a new topic, and 2) when a leaf node of the miniflow is reached, and the DM decides to switch to a different RG on the same topic.
The latter is utilized in the DM's interweaving strategy, wherein a flow-based RG takes turns with an entity-centric or fun fact-based RG in handling a subdialogue on the same topic.

Flow-RG makes it possible for a flow to resume after a few turns handled by a different RG on the same topic. The flow
can simply begin a new miniflow, if there is at least one miniflow that has not yet been visited. Resumption is also possible in the middle of a miniflow, which allows a different RG to chime in for up to two turns (such as EVI answering an on-topic factual question that the flow has no answer prepared for), and then have the miniflow pick up where it left off.

\begin{figure}[h!tb]
    \small
    \begin{tabular}{@{} p{0.425in}|p{2.375in} @{}} \toprule
        ATHENA & Okay, can't wait to get to know a thing or two about you. Valentine's day is on Sunday. Do you have a Valentine?\\
        USER &  yes my wife\\
        ATHENA & Cool! I wish I could meet a socialbot that I could fall in love with; then they'd be my Valentine. Anyway, I've always wanted to go on a vacation and to travel. What's a place that you've always wanted to travel to?\\
        USER & paris sounds great\\
        ATHENA & The Eiffel Tower is a masterpiece of architecture. For me, I think the great wall of china is at the top of my list of travel destinations. It seems like it would be amazing to walk along it. So, Is there a particular reason why you want to go to paris?\\
        USER & the food is really bomb\\
        ATHENA & Delicious food is always a good reason to go somewhere. Hey, I'm curious, Do you prefer flying? or road trips?\\
        USER & I like flying generally\\
        ATHENA & That's an interesting answer. I'd like to go on the road. You can see so much during a roadtrip, it would be cool to watch the world go by. Hmm, Do you like to relax during vacations?\\
        \hline
    \end{tabular}
    \caption{Introduction Flow-RG subdialogue, just before Valentine's day. \label{table:intro_example}}
\end{figure}

\noindent{\bf Introduction RG.}
The Introduction Flow-RG,
which every user experiences,
has a strong effect on the user's overall experience (see Figure~\ref{table:intro_example}).
The Introduction front-loads the conversation with getting-to-know-you content, by learning the user's name and asking icebreaker questions, such as  favorite travel destinations, and weekend activities. The Introduction also brings up relevant current events, such as holidays, 
and gives the user a chance to ask Athena questions. Some of these turns will be the same for most users, e.g., asking for their name. Other content will change based on proximity to significant events in the year or the current day of the week, while some content changes randomly, for example, asking different questions related to vacation preferences. Content related to particular holidays as illustrated in Figure~\ref{table:intro_example} are set up on a calendar and automatically started and stopped. 
The introduction also changes significantly for repeat users to indicate that we remember them, and provide a novel experience.

\subsection{Knowledge Graph-Based RGs}
\label{sec:kg-based-rg}
The goal of the Knowledge Graph-based RGs (KG RGs) is to 
create deep knowledge-grounded conversations, where Athena always has more to say, by traversing relations in the Wikidata knowledge graph.  Athena has four KG RGs covering 
movies, music, sports and TV, with conversations anchored around KG nodes (named  entities). Each topic attempts to continue the conversation by 
either responding with a fact about an entity in context,
or by selecting an entity from a set of fallback entities. When 
the system has either run out of facts on a particular entity, or has been on the same 
entity for a number of turns above a threshold, the RG attempts to traverse one or more relation edges, to a related
entity, to continue the conversation. An example for the TV KG RG is in Figure~\ref{fig:convo_kg}. 
\begin{table}[htb]
    \small
    \begin{tabular}{@{} p{0.425in}|p{2.275in} @{}} \toprule
        ATHENA & Is there a television show that you like? \\
        USER & {\bf Friends} is one of my favorites \\
        &{\scriptsize Entity linking: Friends -> entity: Friends, wiki\_id: Q79784}\\
        ATHENA & Friends is a great show. Why do you like it? \\
        USER & It is really funny \\
        ATHENA & What character do you like the most? \\
        USER & I love Rachel \\
        ATHENA & I really like how Jennifer Aniston plays Rachel 
        Green. Did you know that she won a Primetime Emmy Award
        for Outstanding Lead Actress in a Comedy Series for her 
        role? \\
        USER & No, that's interesting \\
        \hline
    \end{tabular}
    \caption{Conversation using the knowledge graph.       \label{fig:convo_kg}}
\end{table}


Each topic has one to three entity types which the RG can respond about and  each
entity has a set of relations that can be used to generate 
responses. 
Each relation can only be used once for a particular entity, but can be reused when the RG has switched to a new entity. One limitation of the KG RGs is the need to select ``interesting'' relations and write
templates by hand~\cite{moon2019opendialkg}.


\subsection{Entity-Based Indexing RGs}
\label{sec:center-rg}

\begin{table}[htb]
    \begin{scriptsize}
    \begin{tabular}{@{} p{0.35in}|p{0.25in}|p{2.0in} @{}} \toprule
        \bf Topic & \bf \#Facts & \bf Example \\ \hline
        Animals &  90 & I read this surprising fact about koalas. The fingerprints of a koala are so indistinguishable from humans that they have on occasion been confused at a crime scene. Imagine having your fingerprints confused with a koala, how strange!\\ \hline
        Comic Books & 26 & Batman and Robin are the best superhero sidekick team. Once, after Batman and Robin rescued it from a slaughterhouse, DC comics included a Bat-cow.\\ \hline  
         Harry Potter & 21 & Fred and George Weasley were such tricksters. When Fred and George Weasley bewitched snowballs to hit Professor Quirrell's turban, they were unwittingly hitting Voldemort in the face. \\ \hline
        Movies  & 54 & One of my favorite movie series of all time are the James Bond movies. Before signing on as James Bond, Daniel Craig wasn't sure he wanted to play the role.\\ \hline
        Music & 31 & I like to think that when music is played, it makes everyone happier, even animals. Researchers at the University of Leicester found that cows produce 3\% more milk when they listen to relaxing music.\\ \hline
        Nature  & 15 & I like learning more about nature. It's actually really dark in the Amazon Forest! The forest is so thick that only 1\% of sunlight can make it through.\\ \hline
        Video Games  & 20 & Here's a fact I discovered recently about World of Warcraft.  A lot of famous people played World of Warcraft, including Vin Diesel, Mila Kunis and even Robin Williams. Isn't that cool?\\ \hline 
    \end{tabular}
    \end{scriptsize}
    \caption{Fun facts for popular entity-based  RG topics.       \label{table:center_stats}}
\end{table}

Entity-Based Indexing RGs are topical  retrieval-based generators where the focus of the response is on ``fun facts'' for entities in a topic. Table~\ref{table:center_stats} indicates how many fun facts these RGs have for each topic, and provides examples. 

\begin{figure*}[t]
  \centering
  \vspace{-.1in}
  \includegraphics[width=0.95\textwidth]{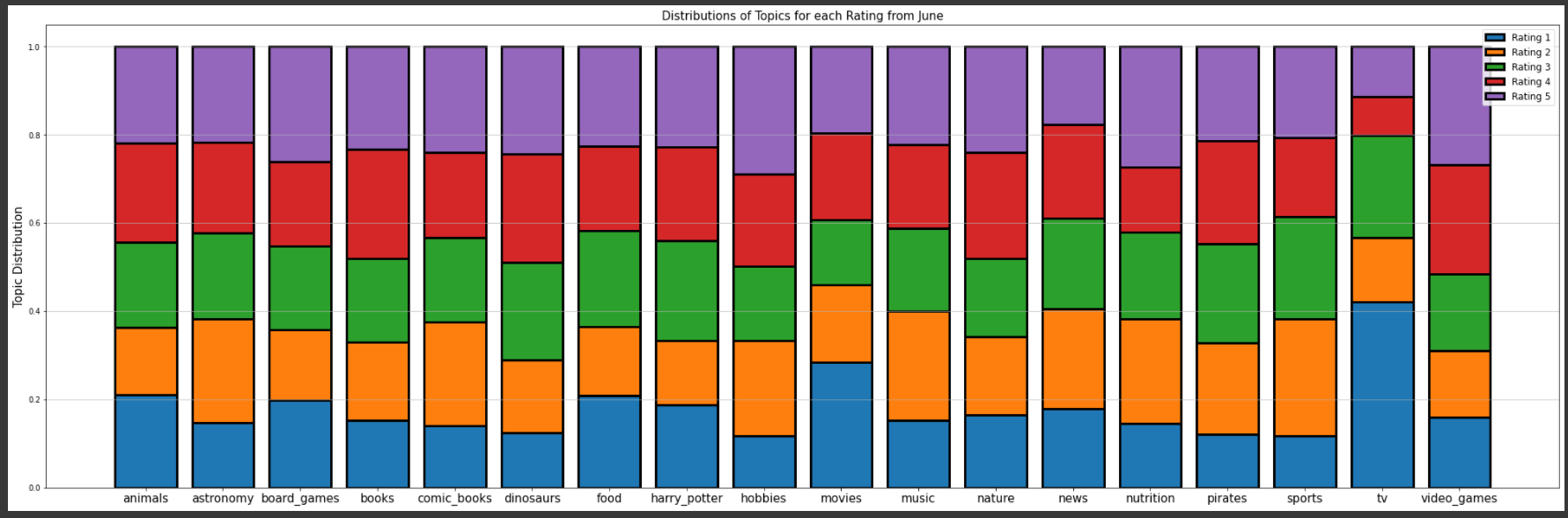}
  \caption{Distribution of ratings for each topic for the period from January 1st to June 16th.}
  \label{fig:ratings-by-topic-distrib}
  \vspace{-.15in}
\end{figure*}

\subsection{Neural NLG RGs}
\label{sec:neural-nlg-rg}

We have also developed and experimented with several different neural NLGs,
including neural NLGs that generate from meaning
representations and are thus topic specific
\cite{juraska2019viggo,harrison2019maximizing, oraby2019curate}.

We also developed a neural NLG that we call Discourse-Driven NRG (DD-NRG)  that generates directly from
the conversation context and can be used for any topic \cite{rajasekaran2020,tosh2020}. We also systematically tested two topic-agnostic neural NLGs provided by Amazon, the PD-NRG \cite{hedayatnia2020policy} and a model called
Topical-NRG that was trained on the Alexa Prize conversations
of all finalists in the 19/20 competition. We found that it was difficult to control the quality of the neural RG outputs and guarantee their coherence, so we only  deployed them to collect experimental data for short periods. We are currently experimenting with methods for  controllable generation for these RGs \cite{reedetal20,harrison2019maximizing,juraskawalker21}.

\section{Evaluation and Analysis}
\label{eval-sec}

The two
criteria that are specified in the Alexa Prize Grand Challenge that systems aim to optimize are length of conversation and user ratings. The Grand Prize will go to a system that achieves conversations of {\bf at least 20 minutes} with {\bf average ratings of 4.0} on a scale of 1 to 5.\footnote{https://www.amazon.science/academic-engagements/alexa-prize-socialbot-grand-challenge-4-finalists-announced}

\begin{figure*}[t!]
  \centering
  \includegraphics[width=0.9\textwidth]{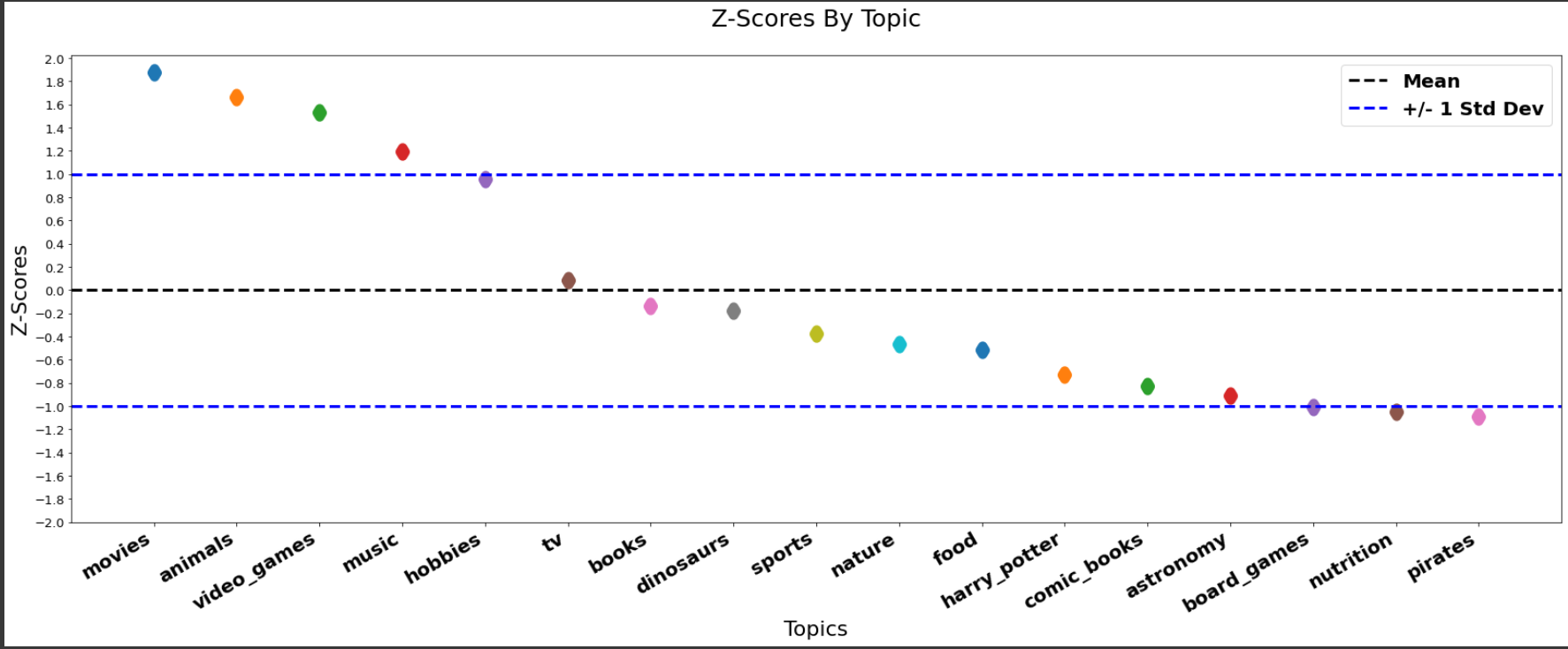}
  \caption{Z-scores for Athena topics for the period from June 1st to June 16th.}
  \label{fig:z-scores-by-topic-june}
  \vspace{-.15in}
\end{figure*}

Over the 4 years our team has been in the competition, we have found that interactions with users are vulnerable to noise due to the competition setup \cite{bowden2019cui,bowden2019slugbot,harrison2020athena}. Users often get into the Alexa Prize skill by accident leading to many conversations of only 1 or 2 turns \cite{shalyminov2018neural}. Surprisingly, even for single turn conversations, some users still provide ratings. To improve our analysis of system performance, we remove these very short conversations from the data. Table~\ref{tab:rating-results} show the ratings,  lengths in turns, and durations, during the semi-finals and the finals. On June 25th, before entering the finals, the average rating across all the systems in the semi-finals was 3.41 and the median duration was 2.12.

\begin{table}[htb]
    \centering
    \small
    \begin{tabular}{p{1.2cm}|p{0.55cm}|p{0.78cm}|p{0.55cm}|p{0.78cm}|p{0.78cm}}
    \hline 
      &\multicolumn{2}{c|}{Ratings} & \multicolumn{2}{c|}{Turns} & {Time}\\
      &Mean & Median & Mean & Median & Median
      \\\hline \hline
       Semifinals  & 3.62 & 4.0 & 17 & 24  & 2.46 \\
       Finals  & 3.71 & 4.0 & 18 & 24 & 2.01
        \\\hline
    \end{tabular}
    \vspace{-.05in}
    \caption{Athena's performance during the semi-finals and the finals for rating, length and duration.}
    \label{tab:rating-results}
    \vspace{-.1in}
\end{table}


Obviously, user's interactions with different RGs and topics affect their
conversations and therefore their ratings. While only about 20\% of users actually provide  ratings, over the course of this year, we collected about 38K conversations with ratings.  The distribution of ratings by topic presence
in conversations from January to June are  in  Figure~\ref{fig:ratings-by-topic-distrib}.
The purple and red bars indicate proportions of the topic that occur in conversations with  ratings of 4 and 5 respectively. 
This suggests that the highest performing topics include animals, comic books, Harry Potter, hobbies, and video games, and that only a few topics are actually performing poorly, such as dinosaurs, news and sports.

However, presence in a conversation is a rather imprecise indicator of topic quality. 
In  order to better understand the contribution
of each topic to Athena's overall ratings, we developed a novel scoring
function that aims to optimize topic selection over the
prizes' user ratings and conversation duration 
criteria.  Thus, our scoring function
gives credit based on the number of utterances in a conversation
that are contributed by each topic in the conversation.
The number of utterances is multiplied by the conversation rating and summed for each topic over all rated conversations. This
sum is then normalized to produce its Z-score. The plot shown
in Figure~\ref{fig:z-scores-by-topic-june} indicates for each topic, how many standard deviations its performance is above or below the
Athena's mean performance for two weeks in June, 2021. This
plot indicates that the topics that contribute most to long dialogues
that are more highly rated include movies, animals, video games, music and hobbies, all of which are 1 or more standard deviations above the mean in performance.  \citet{walkeretal-POD21} discusses the impact of topic selection and dialogue behaviors on user ratings and conversation length. 

We also investigated the impact of personalization on the dialogue  manager with  an A/B study run over 22 days. The A system has the fully personalized dialogue manager, while B utilizes our previously existing  heuristic topic selection strategy. We selected  conversations of 6 turns or more to allow personalization to have an effect and also filter out conversations that never perform topic promotion; thus, we only inspect conversations that differ due to the topic selection strategy. After filtering, personalization occurred in roughly 25\% of conversations. Our results,
shown in Table~\ref{table:personalized_ab_results}, indicate a statistically significant increase in both user rating and conversation length.


\begin{table}[h!]
    \centering
    \small
    \begin{tabular}{c|ccc} 
    \toprule
        & personalized & heuristic & p-value\\
        \hline
        convs & 1066 & 614 & ---\\
        rating & 4.02 & 3.85 & .007\\
        length & 38.07 & 35.07 & .016\\
    \bottomrule
    \end{tabular}
    \caption{Personalized topic selection vs. heuristic topic selection over 22 days.  \label{table:personalized_ab_results}}
    \vspace{-.15in}
\end{table}  

\section{Conclusion}
\label{conc-sec}

This paper describes Athena, and its design and performance in the Alexa Prize Grand Challenge~4. Our system demonstration will
illustrate both the strengths and the weaknesses of Athena
and provide the basis for discussion about the state of the
art of open-domain conversational AI. 

\section*{Acknowledgements}

We would like to thank Christian Benavidez, Yaqing Cao, James Graupera, Colin Harmon, Venkatesh Nagubandi, Meltem Ozcan, Diego Pedro,  Navya Rao, Stephanie Rich, Jasiel Rivera-Trinadad and Aditya Tarde for helping with fun facts,  Wikidata queries and  prosody markup.

%
\bibliographystyle{acl_natbib}
\bibliography{athena}


\appendix

\section{Appendix}
\label{sec:appendix}

\begin{table}[htb!]
    \centering
    \small
        \begin{tabular}{p{2.85in}} 
        \toprule
        \textbf{A:} I'd like to hear your ideas on this. If you could choose one food to live on for your entire life, which food would it be? Why? \\
        \hline
        { \cellcolor[gray]{0.9} \textbf{U:} i think nachos} \\
        \hline
        \textbf{A:} That's an interesting answer! If I was a human, I would eat Sweet potatoes. Sweet potatoes are a really nutritious food. You can eat them in a variety of ways like baked, fries, or twice baked. Anyway, that's my thoughts on the matter. \\
        \hline
        \hline
        \textbf{A:} Okay. I was curious about your opinion on this. If you had the option, would you rather spend time in the mountains, or, at the beach\\
        \hline
        { \cellcolor[gray]{0.9} \textbf{U:} i like the beach more} \\
        \hline
        \textbf{A:} Choosing the beach is a good choice! If I was a human, I would spend time at the ocean - I would walk on the beach, search for shells, and relax in the sun. Anyhow, that's where my minds at, let's move forward. \\
        \bottomrule
    \end{tabular}
    \caption{Two examples of Athena's attempt to increase topical depth by asking different types of playful questions. We crowdsourced 2.5k playful questions, which were then curated for quality. All of this content was also updated to match Athena's self-aware  personality, e.g., \textit{If I was human}. ``A'' denotes Athena and ``U'' the user turns. \label{table:general_conv_example}}
    \vspace{-15cm}
\end{table}

\begin{table}[htb!]
\centering
\small
    \begin{tabular}{p{2.85in}} 
    \toprule
    \textbf{A:} But most people seem to find dinosaurs really interesting. There are dinosaur movies and whole floors of natural history museums dedicated to dinosaurs. Personally, I find it fascinating that modern birds are descended from a particular family of dinosaurs.\\
    \hline
    {\cellcolor[gray]{0.9}  \textbf{U:} hella cool}\\ 
    \hline
    \textbf{A:} Definitely.	I was surprised to hear how much work went into creating the movie Land Before Time. Over 600 background paintings were made for the film. The artists had to create a believable environment in which there was almost no foliage. Most paintings depicted beautiful but barren wastelands.\\
    \hline
    {\cellcolor[gray]{0.9}  \textbf{U:} surprising}\\ 
    \hline
    \textbf{A:} Right?	My favorite dinosaur is the Tyrannosaurus Rex. It's huge and ferocious, but the little arms make it somewhat adorable. What about you? What's your favorite dinosaur?\\
    \hline
    { \cellcolor[gray]{0.9} \textbf{U:} i'm not sure i have to say a t. rex}\\ 
    \hline
    \textbf{A:} Okay. There are lots of movies about dinosaurs. One of the most famous dinosaur movies is Jurassic Park, old but classic. Are there any dinosaur movies that you love?\\
    \hline
    { \cellcolor[gray]{0.9} \textbf{U:} i have to say jurassic park because that's the only one i've seen}\\ 
    \hline
    \textbf{A:} Awesome! That one is well known! I seem to remember people saying the dinosaurs were very vivid\\
    \hline
    { \cellcolor[gray]{0.9} \textbf{U:} i never knew that}\\ 
    \hline
    \textbf{A:} yeah, I love to collect these fun facts! But anyways, I'm interested in your opinion. Would you rather ride on the back of a pterodactyl or brontosaurus as your primary means of transportation?\\
    \hline
    { \cellcolor[gray]{0.9} \textbf{U:} i have to say brontosaurus and i hate heights}\\
    \hline
    \textbf{A:} Choosing a Brontosaurus you say, great pick! I would rather ride a pterodactyl so I could experience flight as well as move around more efficiently. Riding a brontosaurus would be thrilling because of their size but less practical than a pterodactyl. So, okay. Do you know why dinosaurs are extinct?\\
    \hline
    { \cellcolor[gray]{0.9} \textbf{U:}  yeah cause i'm eat your head it right}\\ 
    \hline
    \textbf{A:} Ah nice, you must be super smart. Let me tell you what I heard.The mass extinction of dinosaurs is a scientific mystery, but most hypotheses include an extraterrestrial impact or a massive bout of volcanism. What do you think happened?\\
    \hline
    { \cellcolor[gray]{0.9} \textbf{U:} i'd say aliens}\\ 
    \hline
    \textbf{A:} I'd like to hear your ideas on this. Let's say you are trapped on Jurassic World alone with the dinosaurs, what would you do to survive?\\
    \hline
    { \cellcolor[gray]{0.9} \textbf{U:} i'm not sure i just say find sheltering something and start there}\\ 
    \hline
    \textbf{A:} It's a hard question! I would try to become closer to the younger dinosaurs so that they would grow up with me and feel more comfortable with me. I would also try to help them by finding food and shelter for them. At any rate, Dinosaurs had a lot of different ways to attack other dinosaurs and to defend themselves. To help fight meat eaters such as the Allosaurus or Spinosaurus, many plant eaters had natural weapons at their disposal.\\
    \bottomrule
    \end{tabular}
    \caption{An example conversation about Dinosaurs, one of our more niche topics. We use a variety of different miniflows to increase our topical depth. ``A'' denotes Athena and ``U'' the user turns. \label{table:wyr_hyp_intro_example}}
\end{table} 

\begin{table}[htb!]
    \centering
    \small
        \begin{tabular}{p{2.85in}} 
        \toprule
        \textbf{A:} One thing that really interests me is people's hobbies. Can you give me some info about swimming?  \\
        \hline
        { \cellcolor[gray]{0.9} \textbf{U:} it's pretty fun} \\
        \hline
        \textbf{A:} Right? Sounds totally tubular. Why do you like swimming? \\
        \hline
        { \cellcolor[gray]{0.9} \textbf{U:} it's because i can kind of just feel like i can move around and do a lot of stuff i can't do on land } \\
        \hline
        \textbf{A:} Cool! How did you first get started with swimming? \\
        \hline
        { \cellcolor[gray]{0.9} \textbf{U:} i took swimming lessons } \\
        \hline
        \textbf{A:} I realize you might be totally in love with swimming, but I'm curious. Is there any part of the hobby you don't like? \\
        \hline
        { \cellcolor[gray]{0.9} \textbf{U:} tough lessons } \\
        \hline
        \textbf{A:} Well swimming sounds interesting, thanks for telling me all about your hobby. A lot of people have cool hobbies, but what about me? What kind of hobby do you think I would like? \\
        \hline
        { \cellcolor[gray]{0.9} \textbf{U:} maybe writing } \\
        \hline
        \textbf{A:} Alright. Why do you think writing is a good fit for me? \\
        \hline
        { \cellcolor[gray]{0.9} \textbf{U:} because you like expressing yourself seems like} \\
        \hline
        \textbf{A:} Ok. I see.	Do you need any special equipment or skills to get into writing? \\
        \hline
        { \cellcolor[gray]{0.9} \textbf{U:} just paper and pen and imagination} \\
        \hline
        \textbf{A:}  I see.	Thank you for recommending writing to me, I'll try to find out more about the hobby in the future!\\
        \bottomrule
    \end{tabular}
    \caption{An example of Athena engaging the user about their hobby, swimming, which we previously learned about in the Introduction RG. ``A'' denotes Athena and ``U'' the user turns. \label{table:hobbies_example}}
\end{table} 

\end{document}